\documentclass[letterpaper, 10 pt, conference]{ieeeconf}  % Comment this line out if you need a4paper

\IEEEoverridecommandlockouts                              % This command is only needed if
\usepackage[numbers,sort&compress]{natbib}

\usepackage{graphicx} % for pdf, bitmapped graphics files
\usepackage{verbatim} % For the comment environment.
\usepackage{color}
\usepackage{soul}
\usepackage{subfig}
\usepackage[font=footnotesize]{caption}

\usepackage{amssymb}
\usepackage{amsmath}
\usepackage{algorithm}
\usepackage{algpseudocode}

\usepackage[english]{babel}
\usepackage[T1]{fontenc}
\usepackage[utf8]{inputenc}

\usepackage{xspace}

\usepackage{todonotes}

\usepackage{hyperref}
\usepackage{url}
\usepackage{physics}
\usepackage{siunitx}

\usepackage[nolist,nohyperlinks]{acronym}

\newcommand{\segmatch}{\textit{SegMatch}}

\acrodefplural{CNN}[CNNs]{Convolutional Neural Networks}

\title{\LARGE \bf Learning 3D Segment Descriptors for Place Recognition}

\author{Andrei Cramariuc \and Renaud Dub\'e \and Hannes Sommer \and Roland Siegwart \and Igor Gilitschenski$^{*}$% <-this % stops a space
\thanks{$^{*}$Authors are with the Autonomous Systems Lab, ETH, Zurich
        {\tt\small \{crandrei, rdube, sommerh, rsiegwart, igilitschenski\} @ethz.ch}.}%
        \thanks{
This work was supported by the European Union's Seventh Framework Programme for research, technological development and demonstration under the TRADR project No. FP7-ICT-609763.
}
}

\begin{document}

\begin{acronym}
\acro{CNN}{Convolutional Neural Network}
\acro{SLAM}{Simultaneous Localization and Mapping}
\acro{LiDAR}{Light Detection and Ranging}
\acro{RF}{Random Forest}
\acro{ROC}{Receiver Operating Characteristic}
\acro{GPU}{Graphics Processing Unit}
\acro{SGD}{Stochastic Gradient Descent}
\end{acronym}

\maketitle
\thispagestyle{empty}
\pagestyle{empty}
\bstctlcite{IEEEexample:BSTcontrol}
%%%%%%%%%%%%%%%%%%%%%%%%%%%%%%%%%%%%%%%%%%%%%%%%%%%%%%%%%%%%%%%%%%%%%%%%%%%%%%%%
\begin{abstract}

In the absence of global positioning information, place recognition is a key capability for enabling localization, mapping and navigation in any environment. 
Most place recognition methods rely on images, point clouds, or a combination of both. 
In this work we leverage a segment extraction and matching approach to achieve place recognition in \ac{LiDAR} based 3D point cloud maps. 
One challenge related to this approach is the recognition of segments despite changes in point of view or occlusion. 
We propose using a learning based method in order to reach a higher recall accuracy then previously proposed methods. 
Using \acp{CNN}, which are state-of-the-art classifiers, we propose a new approach to segment recognition based on learned descriptors. 
In this paper we compare the effectiveness of three different structures and training methods for \acp{CNN}. 
We demonstrate through several experiments on real-world data collected in an urban driving scenario that the proposed learning based methods outperform hand-crafted descriptors.
\end{abstract}

%%%%%%%%%%%%%%%%%%%%%%%%%%%%%%%%%%%%%%%%%%%%%%%%%%%%%%%%%%%%%%%%%%%%%%%%%%%%%%%%
\section{INTRODUCTION}

Place recognition is a key capability in \ac{SLAM}. 
In cases where global positioning information is not available, place recognition can be used to compensate for drift and perform loop-closure. 
Amongst other sensors, cameras are commonly used for \ac{SLAM} as they are inexpensive and provide rich visual information. 
However, visual place recognition systems may potentially suffer from large viewpoint or illumination changes~\cite{lowry2016visual}. 
In this work, we consider the alternative of generating 3D point cloud maps using \ac{LiDAR} sensors, which effectively capture the geometry of the environment. 
Particularly, this work focuses on the \segmatch\ method for performing place recognition in 3D point clouds, which is based on comparing segments extracted from a point cloud with previously observed segments~\cite{dube2017segmatch}.

The ability to correctly and robustly match segments is essential in \segmatch. 
In our previous work~\cite{dube2017segmatch}, segment retrieval was performed using descriptors based on the variance along the principal axes of the point cloud segments, followed by a nearest neighbor search in the descriptor space. 
In this work, we evaluate different learning based techniques for achieving this segment description and retrieval. 
For the task of matching segments, the extracted descriptors should ideally be viewpoint invariant, so that the robot is able to recognize a place even if it is approached from a different direction.
Viewpoint invariance is also required because due to drift the robot does not know its true orientation in the environment, which makes perfectly aligning the segments before descriptor extraction impossible. 
Furthermore, point of view invariance is a strong requirement when identifying inter-robot association in multi-robot scenarios. 
As there is often no prior information on the relative transformation between robots, one cannot guarantee that segments will be similarly aligned before descriptor extraction. Invariance to occlusion is also an important aspect of the descriptors, since the robot is not guaranteed to always see entire segments due to occlusion from other objects in the environment. 
Finally, we want the extracted descriptors to be as compact as possible so as to increase performance and reduce the size of the map. 

In this paper we compare the effectiveness of descriptors for 3D point cloud segments produced by three different \ac{CNN} training methods and structures proposed by Parhi \textit{et al.}~\cite{parkhi2015deep} and Hadsell \textit{et al.}~\cite{hadsell2006dimensionality}. 
Through experiments performed on real-world data, taken from the KITTI dataset~\cite{geiger2013vision}, we show that learning based descriptors outperform the descriptors previously used by \segmatch. 
To summarize, this paper presents the following contributions:
\begin{itemize}
\item A new deep-learning based approach to generating descriptors for 3D point cloud segments.
\item Thorough experiments on real-world data demonstrating the effectiveness of learning based descriptors for 3D point cloud segments.
\end{itemize}
The remainder of the paper is structured as follows: 
Section~\ref{sec:related_work} presents a more detailed look at \segmatch\ and related work in generating learning based descriptors. 
Section~\ref{sec:system} describes the three different feature extraction pipelines that were implemented. The generated descriptors are evaluated in Section~\ref{sec:experiments} and Section~\ref{sec:conclusion} finally concludes with a short discussion.

%%%%%%%%%%%%%%%%%%%%%%%%%%%%%%%%%%%%%%%%%%%%%%%%%%%%%%%%%%%%%%%%%%%%%%%%%%%%%%%%
\section{RELATED WORK}
\label{sec:related_work}

\begin{figure*}
\centering
\includegraphics[width=5in]{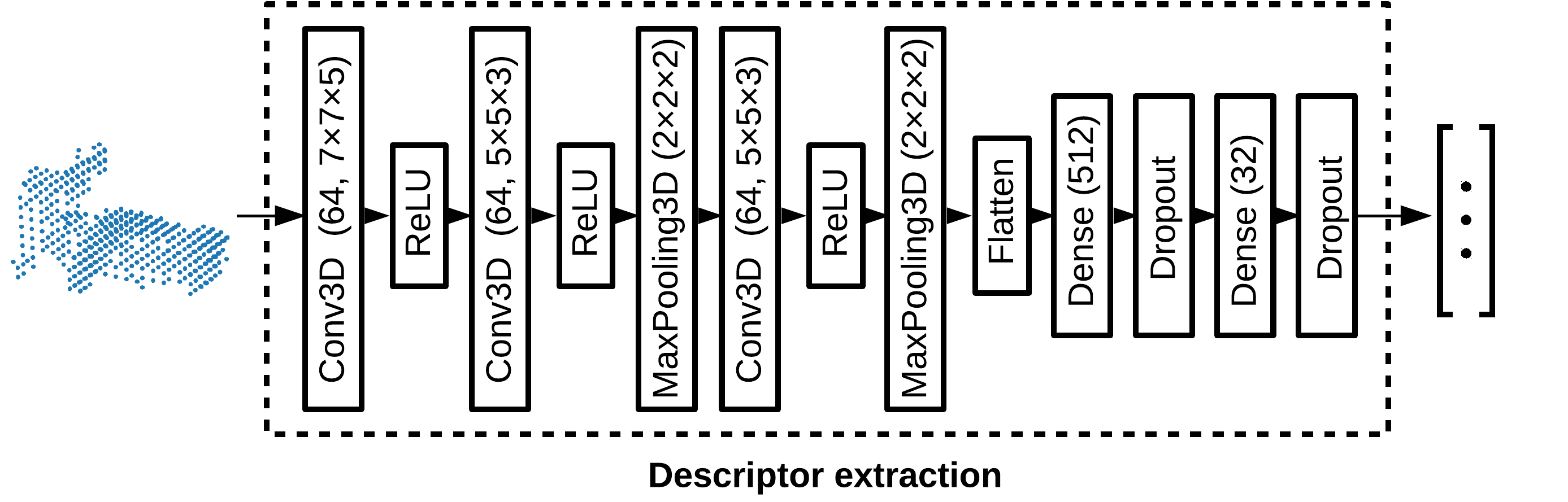}
\caption{A diagram of the descriptor extraction network used in the three proposed methods.}
\label{fig:network_features}
\vspace{-5mm}
\end{figure*}

In recent years, \acp{CNN} have become the state-of-the-art method for generating learning based descriptors, due to their ability to find complex patterns in data~\cite{krizhevsky2012imagenet} having achieved state-of-the-art performance in applications such as object detection in 3D point clouds~\cite{engelcke2017vote3deep}. 
\acp{CNN} have also been successfully used to describe and infer 3D structures~\cite{wu20153d},~\cite{wohlhart2015learning} and generate local key-point descriptors for point clouds~\cite{elbaz2017CVPR}. 
Many of the feature extraction methods described in the literature focus mainly on 2D images, but can be easily applied to 3D point clouds by extending the 2D filters in \acp{CNN} to 3D. 
A neural network can even be designed to directly handle non-discretized point clouds, avoiding problems with aliasing, but requiring accurate information about alignment during training~\cite{qi2016pointnet}.

A common method for generating descriptors with \acp{CNN} is to use the activations of the intermediary layers of the network as a descriptor~\cite{parkhi2015deep}. 
Depending on the networks structure and the training dataset, different types of features can be extracted. 
The disadvantage of this method is that the desired properties of the descriptors are not strictly enforced. 
For face recognition Parkhi \textit{et al.} propose to first train a \ac{CNN} to classify people in the training set and to afterwards use the second to last layer as a descriptor for other faces. 
A more direct approach is that of Hadsell \textit{et al.}~\cite{hadsell2006dimensionality} or Lai \textit{et al.}~\cite{lai2015simultaneous} that propose loss functions which maximize the Euclidean distance between descriptors of dissimilar inputs and minimize it for descriptors of matching inputs. 
While these loss functions guarantee the desired properties, they require a more difficult training process which also affects how easily the model converges. Another popular approach to automatic feature extraction is the use of autoencoders, which are neural networks that can be trained to compress a dataset. 
The compressed representation can then be used as a descriptor~\cite{vincent2008extracting}~\cite{krizhevsky2011using}. 
Autoencoders present an interesting opportunity of accomplishing two tasks at the same time, compression and feature extraction. 
However, this is not guaranteed, since the encoding and the feature extraction part are not directly correlated and can have conflicting goals.

We extend the previous works of Parkhi \textit{et al.}~\cite{parkhi2015deep} and Hadsell \textit{et al.}~\cite{hadsell2006dimensionality} to the 3D domain to create learning based descriptors of point cloud segments. 
This is a new approach to describing point clouds that strikes a balance between local and global features. 
These descriptors are then used in the context of \ac{SLAM} to achieve better performance, as well as a significantly more compressed map that can be more easily stored and shared.

\section{METHOD}
\label{sec:system}

Out of the multiple learning based descriptor extraction methods that were tested, we present the three most successful ones. 
The first two approaches described in this section are similar to the ones used by Parkhi \textit{et al.}~\cite{parkhi2015deep} while the third is based on the contrastive loss function proposed by Hadsell \textit{et al.}~\cite{hadsell2006dimensionality}.

\subsection{Preprocessing}

The raw point clouds from the KITTI dataset are accumulated in a voxel grid segmented using region growing based on Euclidean distances. 
The same method is also used by \segmatch\ as described in our previous work~\cite{dube2017segmatch}. 
Afterwards, the resulting segments are grouped together by looking at successive frames and adding all the segments whose centroids are closer than a manually determined threshold, $d_{same}$, to the same \textit{group}. 
The resulting \textit{groups} therefore combine segments that correspond to the same surface areas in the environment. 
The desired end result of the descriptor extraction is therefore that all the segments belonging to the same \textit{group} should have matching descriptors, while segments that belong to different \textit{groups} should have non-matching descriptors.

The segments are aligned by rotating them so that the robot's position at the time the point cloud was captured would be situated on the positive x-axis. 
This means that the outward part of the surface of the segment that was visible to the robot is always pointing in the same direction. 
This alignment method was chosen since it is simple and robust and allows the robot to align segments in the same way as long as it approaches them approximately along the same path. 
To further increase robustness and make the descriptor extraction process less sensitive to alignment, the dataset is augmented by using multiple copies of the same segment rotated at different angles.

Afterwards, the segments are scaled to fit in to a voxel grid with a fixed size and centered inside the grid. 
A larger voxel grid allows for more details, but also requires more time and computational power to process. For segments that are larger than the size of the grid, the outlying points are discarded. 
The voxelized segments are then normalized by removing the mean and dividing by the standard deviation separately for each voxel. 
This causes gradients to act uniformly and in practice reduces convergence time while sometimes also increasing the accuracy.

Finally, segments that are too similar after the preprocessing are removed. 
Similar segments occur often, since between two successive frames many segments that are not located at the boundaries of the \ac{LiDAR} scanner remain the same. 
The dataset is filtered by comparing the Hamming distance between all pairs of voxelized segments inside the same \textit{group} and removing one of the segments in a pair if the Hamming distance is below a threshold, $th_{H}$.

\subsection{Feature extraction network}

The same descriptor extraction part of the \ac{CNN} is used in all three methods. 
This way results are easily comparable and the structure of the descriptor extraction networks needs to be optimized only once. 
A well performing structure of the descriptor extraction \ac{CNN} is presented in Figure~\ref{fig:network_features} and was found by grid searching through different parameters. 
Different depths and sizes for the the layers and filters were tested, while keeping the network small enough so that it would be feasible to run on a mobile robotic platform. 
The only difference between the utilised descriptor extraction networks is the amount of dropout in the final layers, which was tuned separately for each method to ensure a correct regularization.

\subsection{Group based classification}
\label{sec:groups}

The first approach is to train a \ac{CNN} to classify segments based on the \textit{group} to which they belong, meaning that each \textit{group} represents a class.
The layer before the classification layer can be used as a descriptor~\cite{parkhi2015deep}, as presented in Figure~\ref{fig:groups}.
Due to the large number of classes and the small dimensionality of the descriptor the network is forced to focus on descriptive and not specific features.
The proximity in the Euclidean space between descriptors of segments in the same \textit{group} is loosely enforced by the simplicity of the classification layer.
The probability that a segment belongs to a class is proportional to the dot product between the weights for that class and the descriptor.
Therefore, descriptors that have a small Euclidean distance, will also most likely be classified as belonging to the same \textit{group}.
The correlation between similarity and the Euclidean distance between descriptors is important as it can be used to quickly generate candidate matches.
If needed a secondary classifier can be trained to predict if pairs of descriptors match or not. 
The network is trained using \ac{SGD} to minimize the categorical cross-entropy. 
Additionally the dataset is filtered removing \textit{groups} with too few samples.

\subsection{Siamese convolutional neural network}
\label{sec:siamese}

\begin{figure}
\centering
  \mbox{\subfloat[]{\includegraphics[width=3.4in]{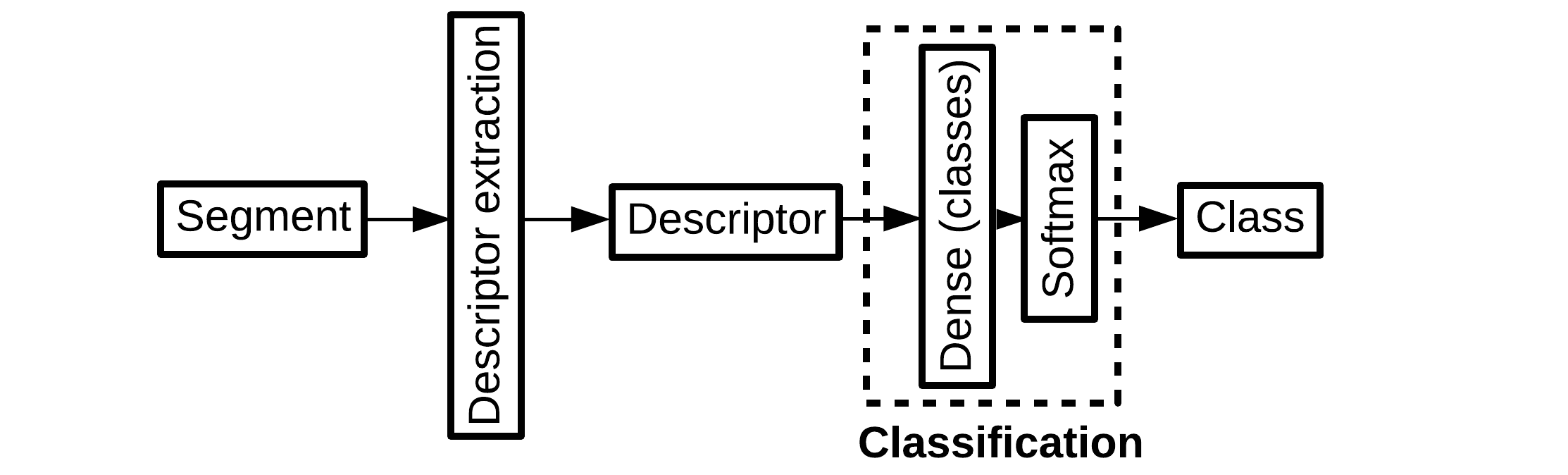} \label{fig:groups}}}
  \mbox{\subfloat[]{\includegraphics[width=3.4in]{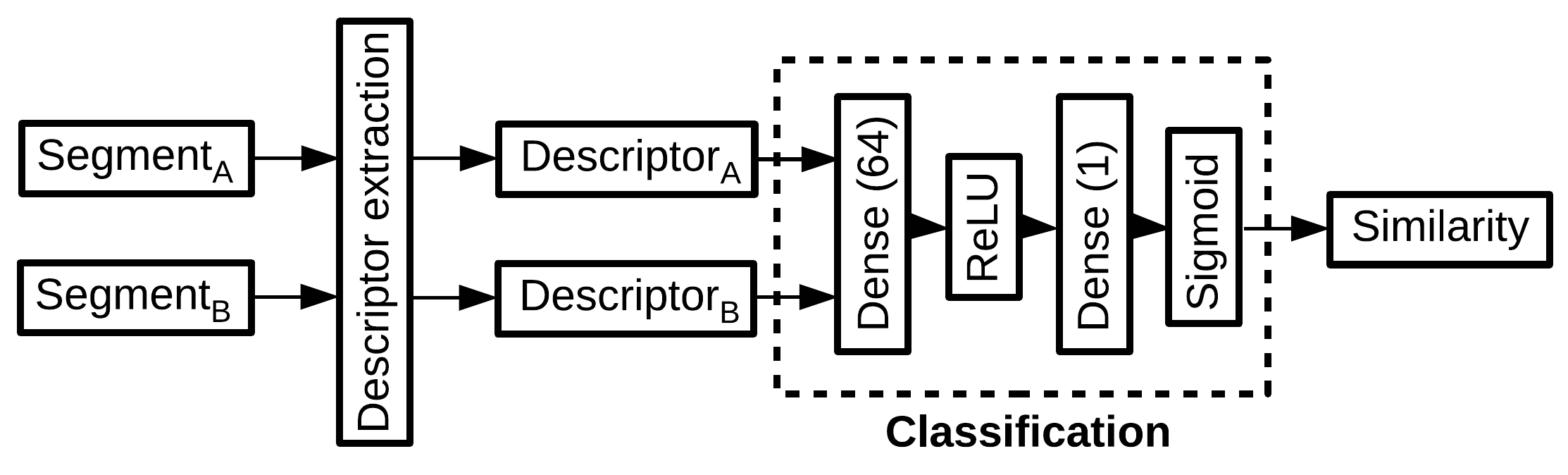} \label{fig:siamese}}}
  \mbox{\subfloat[]{\includegraphics[width=3.4in]{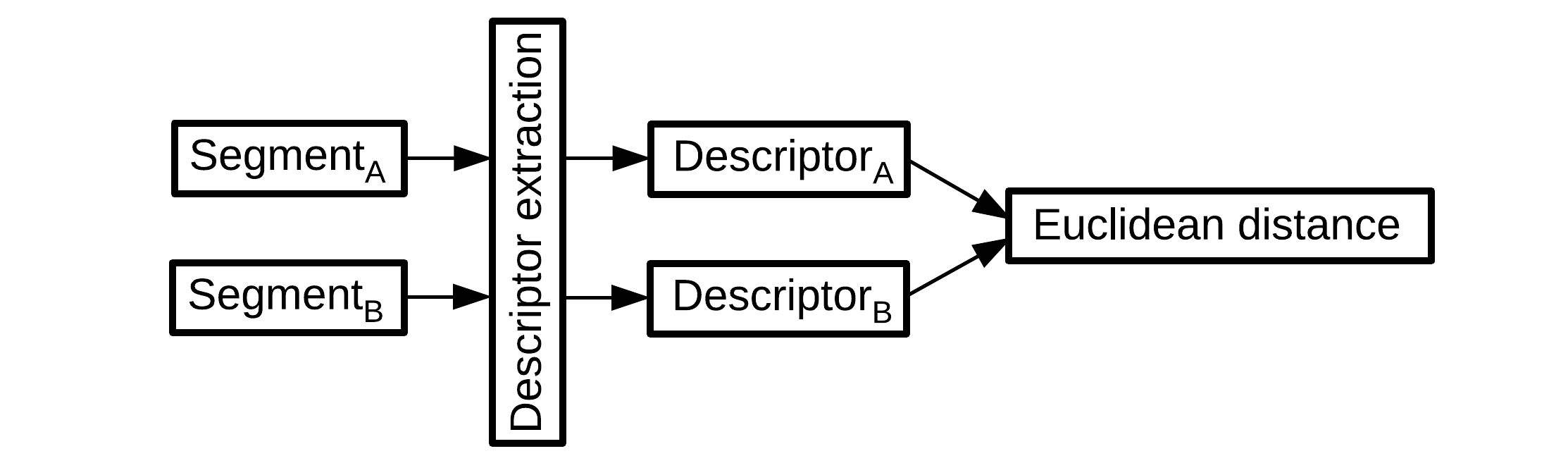} \label{fig:contrastive}}}
  \caption{Diagram of the network structure of the (a) \textit{group} based \ac{CNN} (b) Siamese \ac{CNN}, and (c) descriptor extraction \ac{CNN} trained using contrastive loss.}
  \vspace{-5mm}
\end{figure}

The second approach is to train a Siamese convolutional neural network~\cite{baldi1993neural}. 
A Siamese network takes two inputs that are given to two distinct \acp{CNN}. 
Afterwards, the outputs of these two networks are combined and passed through a third network that produces the final output. 
In this case branches of the Siamese network are two identical descriptor extraction networks, which share the same weights.
The final network detailed in Figure~\ref{fig:siamese}, given the two descriptors, outputs the probability that the two input segments match. 
The advantage of using a Siamese network is that it allows training the feature extraction simultaneously with the classification, as opposed to the two-stage method described in the previous section. 
During inference, the feature extraction part and the classifier can be used independently, which can significantly boost performance when the same feature vector is tested multiple times. 
For training \ac{SGD} is used to minimized the binary cross-entropy of the network. 
The training and testing pairs are chosen at random, while avoiding duplicate pairs and maintaining a one-to-one ratio between positive and negative pairs.

\subsection{Descriptor extraction using contrastive loss}
\label{sec:contrastive}

The third approach is to train a classifier using contrastive loss~\cite{hadsell2006dimensionality}. 
This aims to minimize the Euclidean distance between the feature vectors of matching segment pairs while maximizing it for non-matching pairs. 
The loss function is applied directly to the resulting descriptors, without performing any classification, as illustrated in Figure~\ref{fig:contrastive}. 
The contrastive loss function is defined as
\begin{equation}
\text{L}(y, f_\text{a}, f_\text{b}) = y*|| f_\text{a} - f_\text{b} ||_2^2 + (1-y) * \max(0, m - || f_\text{a} - f_\text{b} ||_2^2)
\end{equation}
where $f_\text{a}$ and $f_\text{b}$ are the two feature vectors outputted by the network, and $y$ is the ground truth whether the feature vectors should match or not. 
The $m$ parameter is the distance margin that the loss function attempts to enforce between positive and negative pairs. 
The utilized training regime that is to choose hard pairs, which are the pairs of segments for which the Euclidean distance between their descriptors is lowest, but the segments do not match and vice versa. 
The hard pairs are subsampled as suggested in~\cite{lai2015simultaneous},~\cite{schroff2015facenet} to avoid overly focusing on only specific cases. 
For increasing performance and avoiding getting stuck in local minima, the hard pairs are recalculated only at the end of each training epoch.

%%%%%%%%%%%%%%%%%%%%%%%%%%%%%%%%%%%%%%%%%%%%%%%%%%%%%%%%%%%%%%%%%%%%%%%%%%%%%%%%
\section{EXPERIMENTS}
\label{sec:experiments}

\begin{figure*}
\centering
  \captionsetup[subfigure]{oneside,margin={0.3in,0in}}
  \mbox{\subfloat[]{\includegraphics[width=2.25in]{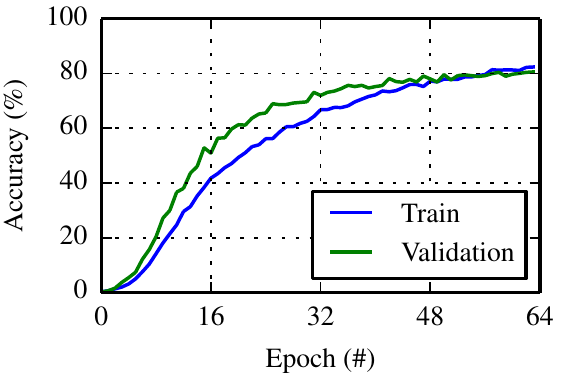} \label{fig:groups_acc}}}
  \mbox{\subfloat[]{\includegraphics[width=2.25in]{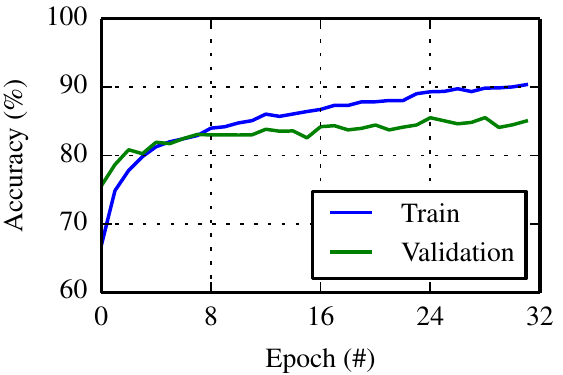} \label{fig:siamese_acc}}}
  \mbox{\subfloat[]{\includegraphics[width=2.25in]{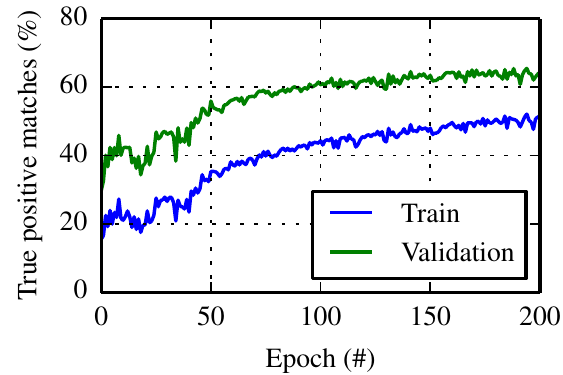} \label{fig:contrastive_acc}}}
  \caption{(a) Classification accuracy of the \textit{group} based network during training. 
  	(b) Classification accuracy of the siamese network during training. 
  	(c) The accuracy with which correct candidate matches were generated while training the descriptor extraction network using contrastive loss.}
  \vspace{-5mm}
\end{figure*}

The different descriptor extraction models were trained and tested on real world data from the KITTI dataset. 
The raw data consisted of the output of a Velodyne laser scanner mounted on an autonomous driving platform, that drove through two residential areas in the city of Karlsruhe. 
The segmentation of the dataset was achieved using the same \ac{LiDAR}-based \ac{SLAM} system as \segmatch~\cite{dube2017multirobot}. 
The threshold $d_{same}$ was manually set low enough to \SI{1.5}{m}, so as to avoid false positives. After segmentation and preprocessing the two drives, numbered 18 and 27, contained approximately 12000 and 16000 segments, which formed 2000 and 3000 \textit{groups} respectively. 
The first drive was split into a training and validation set, while the second drive was used as a testing set.

The segments were fit in to a $38\times38\times18$ voxel grid, the size of which was determined empirically so as to strike a good balance between the ability to differentiate segments and the size of the required network. 
The voxel size of approximately \SI{0.2}{m} in all dimensions, was chosen so that approximately 95\% of the segments would fit entirely into the voxel grid. 
The mean and standard deviation were computed on the training set, and then used to normalized the segments in the entire dataset. 
The $th_H$ threshold was manually set to 50, meaning that segments with less than 50 points different than any other segment in the same \textit{group} are considered duplicates and filtered out.

The performance of the different methods was evaluated based on how well they discriminate between matching and non-matching pairs of segments, as well as based on how many correct candidate matches are obtained by looking at the closest neighbor in the Euclidean space of the descriptors. The first metric is indicative of how well the model understands the underlying patterns in the dataset. The second metric tells us how well imposed the desired properties of the descriptors are and therefore how useful the descriptors are for recognizing places.

\subsection{Training the models}

For the \textit{group} based model detailed in Section~\ref{sec:groups}, \textit{groups} with too few segments were first removed from the training set. 
The resulting training set had approximately 6600 segments and 500 \textit{groups}, which are equivalent to the classes of the \textit{group} based network. 
The classification accuracy of the model at each epoch during training is presented in Figure~\ref{fig:groups_acc}. 
Both, the training and validation accuracy match, and the \textit{group} based model reaches a final classification accuracy of approximately 80\%.

The training results for the Siamese \ac{CNN} detailed in Section~\ref{sec:siamese} are presented in Figure~\ref{fig:siamese_acc}. 
The model quickly learns during the first few epoch, after which the validation accuracy remains almost constant, while the training accuracy continues to increase. 
It is possible that the model quickly starts overfitting on the significantly smaller amount of positive pairs that can be formed from the training dataset, while it is still learning from the larger amount of negative pairs. 
Adding more regularization reduces the amount of overfitting, but does not affect the final validation accuracy, which is about 85\%.

When training the feature extraction \ac{CNN} using contrastive loss as detailed in Section~\ref{sec:contrastive}, the loss function itself is not very telling of the performance of the model as the training set changes at the end of each epoch. 
The performance of the network during training can be observed by analysing the distribution of Euclidean distances of the positive and negative pairs. 
Training with hard pairs decreases the variance of both distributions, which causes less outliers and therefore increases the chance of finding true positive matches when looking at only the closest neighbor in the Euclidean space of a descriptor. 
To asses that the training process is successful, at the end of each epoch for each segment, we look for its corresponding pair based on the shortest Euclidean distance of the descriptors, and then calculate how many correct pairings we obtain. 
The percentage of correct pairings for each epoch is presented in Figure~\ref{fig:contrastive_acc}. The offset between accuracy on the training and validation sets is due to the different number of samples. 
Larger datasets have more possible negative pairings and therefore an increased chance of finding a negative pair with a low Euclidean distance. 
The highest accuracy on the training set is 52\% and on the validation set 65\%.

All the \acp{CNN} were trained on a Nvidia GeForce GTX 980 \ac{GPU}. 
The fastest to train is the \textit{group} based network, which takes only approximately an hour. 
The training set of the other networks is much larger, since it also contains the segments that belong to small \textit{groups} and because the possible number of segment-pairs is far larger than the number of segments. 
The Siamese network requires approximately six hours to converge and the feature extraction network trained using contrastive loss requires a day to converge.

\begin{figure}
\centering
\includegraphics[]{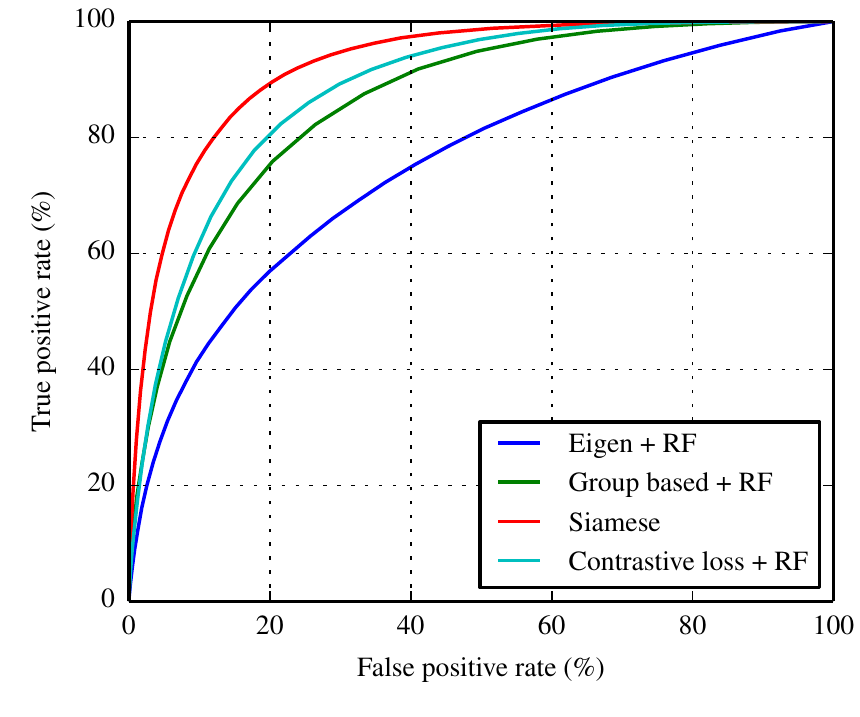}
\caption{The receiver operating characteristic curve for when the different descriptors are used to classify pairs of segments as matching or not.}
\label{fig:roc}
\end{figure}

\begin{figure}
\centering
\includegraphics[]{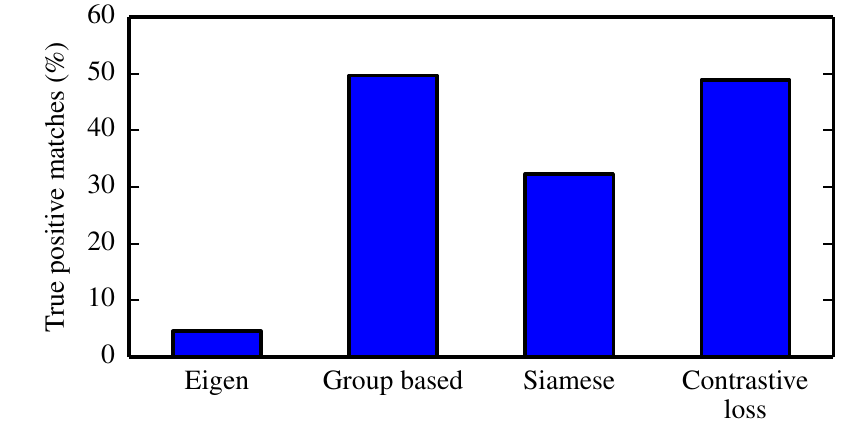}
\caption{The accuracy with which each type of descriptor can be used to generate correct candidate matches.}
\label{fig:comparison}
\vspace{-5mm}
\end{figure}

\subsection{Comparing the different classifiers}

First, we compare the general accuracy of the different descriptor extraction methods, by using positive and negative pairs of segments from the test set. 
For all the other methods except the Siamese network we train a \ac{RF} as a secondary classifier, which takes as input the two concatenated descriptors and outputs the probability that they match. 
Similar results can be obtained using a neural network, a \ac{RF} was used for efficiency purposes. 
As a reference, we include in our analysis the eigenvalue based descriptors currently used by \segmatch. 
The \ac{ROC} curves of the different classifiers are presented in Figure~\ref{fig:roc}. 
The best accuracy is achieved by the Siamese network, which together with the other \ac{CNN} based approaches outperforms the baseline. 
The overall classification accuracy of the Siamese network was approximately 80\%.

The second comparison we make is to see how good the different methods are at generating candidate matches based on the closest neighbor in the Euclidean descriptor-space. 
For each segment in the test set we find a candidate match and check how many of the pairings contain segments that belong to the same \textit{group}. 
The results for all the methods are presented in Figure~\ref{fig:comparison}. 
The \textit{group} based classifier together with the feature extraction network trained using contrastive loss perform the best. 
They both generate around 50\% positive matches in a dataset of about 16000 segments, while the Siamese network has only around 30\% positive matches. 
All the aforementioned methods far outperform the eigenvalue based approach, which only generates approximately 6\% true positive matches. 
The \textit{group} based network is more than 20 times faster to train, but the desired features in the descriptors are less strictly imposed than when training with contrastive loss.

The disadvantage of the \ac{CNN} based approach is the significantly increased requirement in computational power. 
As our main focus was to prove that learning based descriptors can correctly model 3D point cloud segments the descriptor extraction \acp{CNN} were designed to achieve the best performance disregarding throughput. 
On the Nvidia GeForce GTX 980 GPU the descriptor extraction network has an approximate throughput of 680 segments per second. 
In contrast on an Intel i7-6700K CPU the same network only has a throughput of 12 segments per second. 
To achieve the desired real-time throughput required by \segmatch\ the network was reduced in size, which also caused a decrease in accuracy. 
The optimized descriptor extraction \ac{CNN} achieves a 30 times higher throughput, of 340 segments per second, while the amount of true positive matches generated was reduced to approximately half.

\addtolength{\textheight}{-13cm}   % This command serves to balance the column lengths

%%%%%%%%%%%%%%%%%%%%%%%%%%%%%%%%%%%%%%%%%%%%%%%%%%%%%%%%%%%%%%%%%%%%%%%%%%%%%%%%
\section{CONCLUSION}
\label{sec:conclusion}

For place recognition in 3D \ac{LiDAR} maps based on segment matching, using well designed descriptors allows for fast recall, a small map size, and invariance to viewpoint changes. 
Three different \ac{CNN} structures and training methods for generating descriptors for 3D point cloud segments were implemented and tested on real world data. 
These methods were compared against the eigenvalue based descriptor extraction algorithm currently used by \segmatch. 
Firstly, the methods were compared based on how well descriptors could be used to tell if two segments match or not. 
Here the Siamese network performed the best, achieving an overall accuracy of more than 80\%. 
Secondly, we calculated the accuracy with which candidate matches can be generated based on a descriptors closest neighbor in the Euclidean space. 
Of the aforementioned methods, here the \textit{group} based feature extraction network and the feature extraction network, trained using contrastive loss, performed the best. 
Both methods were able to generate approximately 50\% positive matches on a dataset of more than 16000 segments. 
All the \ac{CNN} based methods outperformed the eigenvalue based approach by a significant margin, as it was only able to generate 6\% positive matches. 
Furthermore we showed that the network can be optimized to achieve the required throughput for real-time operation while still maintaining a higher accuracy. 
This proves that the \acp{CNN} were able to correctly model the dataset of 3D point cloud segments and successfully generate descriptors with the desired properties.

%%%%%%%%%%%%%%%%%%%%%%%%%%%%%%%%%%%%%%%%%%%%%%%%%%%%%%%%%%%%%%%%%%%%%%%%%%%%%%%%

%\section*{ACKNOWLEDGMENTS}

\small
\bibliographystyle{IEEEtranN}
\bibliography{bibliography/IEEEabrv,bibliography/iros17}

\end{document}